\documentclass[10pt,twocolumn,letterpaper]{article}
\pdfoutput=1

\usepackage{3dv}
\usepackage{times}
\usepackage{epsfig}
\usepackage{graphicx}
\usepackage{amsmath}
\usepackage{amssymb}
\usepackage{comment}
\usepackage{enumitem}

\threedvfinalcopy 

\def\threedvPaperID{199} 

\ifthreedvfinal\fi
\begin{document}

\title{Progression Modelling for Online and Early Gesture Detection}

\author{Vikram Gupta\\
Mercedes-Benz R\&D India, Bangalore\\
{\tt\small vikram.gupta@daimler.com}
\and
Sai Kumar Dwivedi\\
Mercedes-Benz R\&D India, Bangalore\\
{\tt\small saikumar.dwivedi@daimler.com}
\and
Rishabh Dabral\\
Indian Institute of Technology, Bombay\\
{\tt\small rdabral@cse.iitb.ac.in}
\and
Arjun Jain\\
Indian Institute of Technology, Bombay \\Axogyan AI, Bangalore \\
{\tt\small arjunjain@gmail.com}
}

\newcommand{\sai}[1]{\textcolor{red}{#1}}

\maketitle

\begin{abstract}
   Online and Early detection of gestures is crucial for building touchless gesture based interfaces. These interfaces should operate on a stream of video frames instead of the complete video and detect the presence of gestures at an earlier stage than post-completion for providing real time user experience. To achieve this, it is important to recognize the progression of the gesture across different stages so that appropriate responses can be triggered on reaching the desired execution stage. To address this, we propose a simple yet effective multi-task learning framework which models the progression of the gesture along with frame level recognition. The proposed framework recognizes the gestures at an early stage with high precision and also achieves state-of-the-art recognition accuracy of 87.8\% which is closer to human accuracy of 88.4\% on the NVIDIA gesture dataset in the \textit{offline} configuration and advances the state-of-the-art by more than 4\%. We also introduce tightly segmented annotations for the NVIDIA gesture dataset and setup a strong baseline for gesture localization for this dataset. We also evaluate our framework on the Montalbano dataset and report competitive results.

\end{abstract}

\section{Introduction}
Gestures are, arguably, the oldest form of human communication. Unlike spoken language, they are easier to learn for people belonging to different demographics. Gestures are crucial for people with hearing or speech impairments and are also effective in noisy conditions. These properties make gestures suitable for designing universal and robust interfaces for touchless human computer interaction (HCI). These interfaces can be used to design intelligent car interiors to enable convenient user interaction with multimedia systems, reading lights, sunroof etc. without distracting the driver. Such interfaces are also suitable for designing immersive gaming and augmented reality experiences. 

While gestures are intuitive and easy to learn, gesture recognition is a challenging task as there are different ways and velocities at which people can perform gestures. Variations in the ambient conditions like lighting, background, occlusion further increase the complexity. Gesture recognition systems designed for interactive applications should also address:~\textit{online operation} and~\textit{early prediction}. These systems should work in an online setting, where the gesture recognition is done on an incoming stream of video frames instead of the complete video. They should also respond in real time as a response time of more than 100 ms degrades the user experience \cite{mitra2007gesture},~\cite{card1991information}. To address this, it is important to recognize and predict the gesture earlier than its completion so that appropriate response can be triggered in real time. Different stages of the gesture can be used to trigger early prediction, but it is difficult to define this optimally at the training time as it is guided both by the domain requirements as well as the early prediction vs precision trade-off characteristics of the model. We propose that modelling the complete temporal progression of the gesture along with recognizing the gesture at frame level addresses the above problems and helps to design interactive gesture recognition systems. Our method can be used across different type of gestures as it does not depend on explicit sub-gesture level annotations or information about the structure of the gestures. 

\begin{figure}[t]
\centering
\includegraphics[width=\columnwidth]{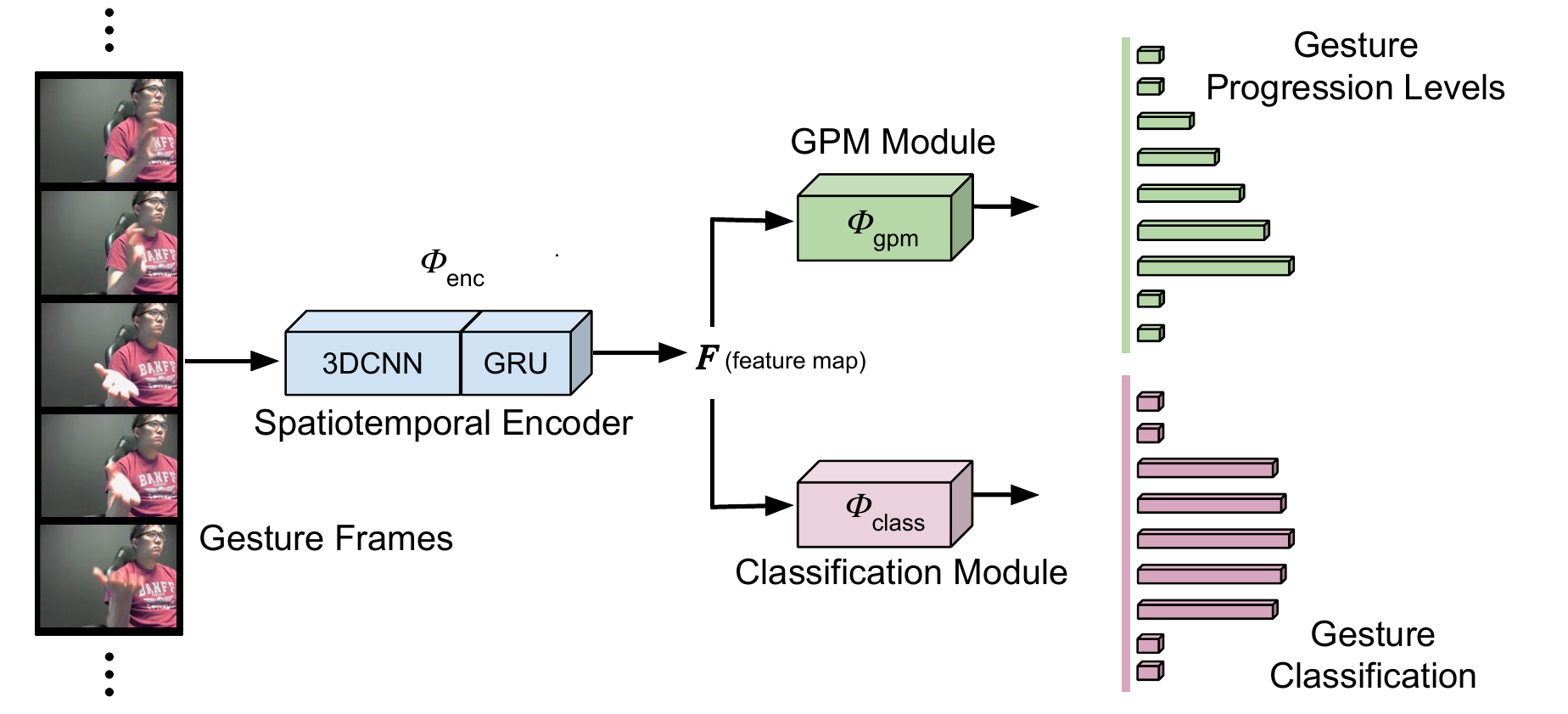}
\caption{Overall schema of our framework. We use a branched architecture with 3DCNN and GRU as spatiotemporal feature extractor and estimate the gesture progression levels and gesture category by GPM and Classification module respectively.}
\label{fig:architecture}
\end{figure}

A lot of research has been conducted in the field of gesture recognition. However, most of these approaches do not address online operation and early prediction. Gesture recognition approaches proposed by Narayana~\etal~\cite{focus}, Miao~\etal~\cite{miao2017multimodal} operate in an offline setting where the recognition is done after the gesture has finished. Localization based approaches as explored by Pigou~\etal~\cite{pigou2018beyond} perform online frame level gesture classification but can not be used for early prediction as the gesture progression is not modelled. This makes it difficult to decide when to trigger the response. Simple heuristics like using the number of frames as the threshold for triggering prediction do not work well as gestures are of different duration. "Clockwise" gesture in NVIDIA gesture dataset \cite{pavloctc} has a mean duration of 0.8 seconds while "one finger tap" has 0.4 seconds. Even the same gesture can be performed at varying speeds. Duration of "swipe right" gesture in the dataset ranges from 0.35 to 1 second.

Molchanov~\etal~\cite{pavloctc} explored early gesture detection using connectionist temporal classification (CTC)~\cite{ctc}. CTC loss function enables gesture detection without requiring frame level annotations which makes it useful as annotation is time consuming and expensive. However, the system learns to detect only a segment of the gesture instead of the complete gesture. Moreover, the location or duration of this segment can not be changed, which makes it difficult to adapt the gesture predictions to meet the domain requirements.

In this work, we propose a simple yet effective multitask learning framework to address \textit{online operation} and \textit{early prediction}. Our framework consists of two sub-modules which operate simultaneously on every frame: classification module and gesture progression modelling (GPM) module. The classification module learns to recognize the category of the gesture and the GPM module models the progression level of the gesture. GPM allows the system to perform early prediction and also provides the flexibility to configure the early prediction stages of the gestures even after the model has been trained. This flexibility is desirable as it saves the time and efforts for training separate models for different early prediction stages. The proposed framework is generic and works well in both \textit{online} and \textit{offline} settings. 

We performed extensive quantitative and qualitative experiments on the NVIDIA and Montalbano gesture datasets to demonstrate that our framework is able to recognize the gestures early with high accuracy and also performs simultaneous classification and detection of gestures. The experiments show that the GPM branch models the progression of the gesture and also improves the offline detection accuracy. We outperform the state-of-the-art results in offline gesture detection on the NVIDIA gesture dataset and report competitive results on the Montalbano dataset. 

Since our goal is to recognize the gesture category and progression accurately at frame level granularity, start and end point annotations are required. However, the NVIDIA dataset annotations are loosely segmented containing background frames also. To bridge this gap, we re-annotated the NVIDIA dataset. The new annotations will be made public. With these tight annotations, we also setup a localization baseline over this dataset. In summary, the key contributions of our paper are:
\begin{itemize}[noitemsep]
    \item A novel multitask framework for online and early gesture recognition. The framework also demonstrates competitive performance on offline gesture classification and localization.
    \item A new state-of-the-art result on offline gesture detection on the NVIDIA dataset which is closer to the human accuracy of 88.4\%~\cite{pavloctc}.
    \item Strongly segmented gesture annotations for NVIDIA dataset for future research and a new localization baseline on the NVIDIA dataset.
\end{itemize}

\section{Related Work}

Classification of dynamic hand gestures has been explored extensively by the research community~\cite{asadi2017survey}, ~\cite{cheng2016survey},~\cite{rautaray2015vision}. Majority of the approaches today leverage deep 2D/3D convolutional neural networks (CNN) and recurrent units for modelling spatiotemporal information and Support Vector Machines(SVM)/Neural Network(NN) based classifiers for the classification. Miao~\etal~\cite{miao2017multimodal} used a combination of residual and C3D model for extracting features from the multi-modal gesture data. The extracted features of the different modalities are fused with a canonical correlation analysis and classified using a SVM. Narayana~\etal~\cite{focus} reported state-of-the-art gesture recognition results on ChaLearn IsoGD~\cite{chalearnLatestDataset} and NVIDIA dataset~\cite{pavloctc} by introducing multiple spatial channels for each modality. The complete image and the crops corresponding to the left and right hand are treated as separate channels so that the model can focus on the hands along with the complete image. The features of these channels are fused using a sparse network to avoid overfitting and classified using a SVM. While these approaches demonstrate promising results, they are mainly designed for offline gesture classification where the task is to classify the gestures after completion. 

Gesture localization approaches that perform frame level classification without processing the whole gesture are more suitable for online gesture classification. Pigou~\etal~\cite{pigou2018beyond} explored frame level gesture classification on the Montalbano dataset~\cite{chalearn2014dataset} using a deep neural network consisting of temporal convolutions and recurrent units. 
Neverova~\etal~\cite{neverova2016moddrop} demonstrated gesture localization by treating classification and localization as two different tasks. The frame level classifications are post processed by a localization module which is a binary classifier to distinguish between gesture and no-gesture. On similar lines, Wang~\etal~\cite{wang2016large} explore a two stage process where they localize the presence of the gesture based on the motion information and then create a depth motion map for the identified segment for classification. However, this two stage process makes this method unsuitable for online gesture classification. While these approaches output predictions for every frame without observing the whole gesture, they do not model the progression of the gesture which makes it difficult to determine the frame at which the appropriate response should be fired. Due to this limitation, the above methods can not be directly used for early gesture recognition.

Molchanov~\etal~\cite{pavloctc} studied early gesture detection by using CTC as the loss function to detect the nucleus of the gesture. However, their method does not detect the complete gesture and can not be used for prediction at any other stages apart from the nucleus.

Early event detection has been explored in the literature but with limited focus as compared to classification and detection. Hoai~\etal~\cite{early_prediction_2} used a Structured Output SVM (SOSVM) to identify the events from partial observations. Temporal segments encompassing the event partially or completely are used as positive samples and remaining segments are used as negative samples to train the SOSVM to distinguish between the background and the event. The authors also constrain the classifier to output higher confidence score as it observes more of the event. Ma~\etal~\cite{early_prediction_3} propose a new ranking loss function to encourage the model to become more confident as the activity progresses. The loss function constrains the model to output monotonically non-decreasing detection score for the correct category or the margin between the correct category and the category with highest probability. Aliakbarian~\etal~\cite{early_prediction_4} also propose a new loss function which applies monotonically increasing penalty to the model for incorrect predictions along with cross entropy loss to discourage the model from generating false positives as it observes more of the activity. Although, these approaches help the model in predicting the activity at an early stage, they do not model the progression of the activity separately. 

Inspired by the above approaches, we propose a novel multitask framework which addresses early and online gesture classification by modelling gesture classification and progression separately into two different branches. Explicit modelling of the gesture progression provides the flexibility to specify the gesture trigger points even after training. We demonstrate that our framework performs better early gesture prediction as well as offline classification and localization.

\begin{figure*}[h]
\begin{center}
\includegraphics[width=\linewidth]{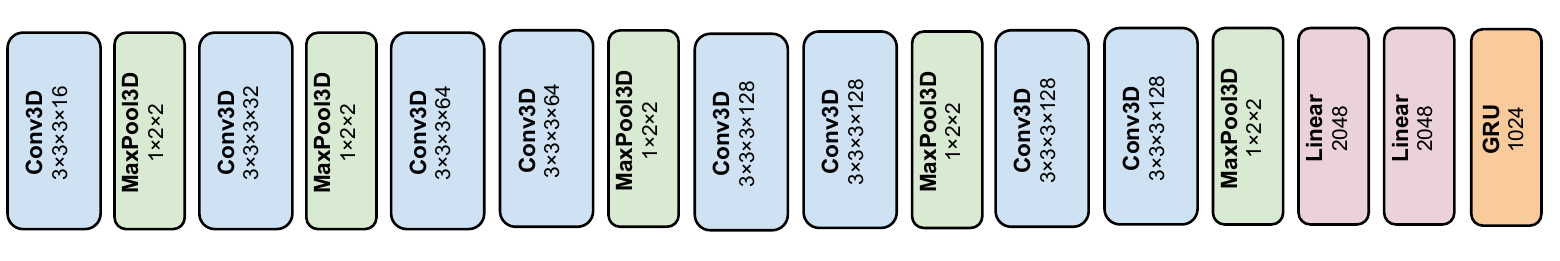}
\end{center}
\caption{Architecture of the spatiotemporal encoder. All the 3D convolution kernels are 3 x 3 x 3 with the denoted number of filter maps and are followed by batch normalization. All the pooling layers preserve the temporal dimension and have the kernel size of 1 x 2 x 2. Both the linear layers have 2048 units and the  Gated recurrent unit (GRU) has 1024 units. ReLU is used as the activation function.}
\label{fig:spatio_temporal_architecture}
\end{figure*}

\section{Proposed Method}
Our architecture consists of three core components: Spatiotemporal Encoder ($\Phi_{enc}$), Gesture Progression Modelling (GPM) module ($\Phi_{gpm}$) and Classification module ($\Phi_{class}$) as shown in Figure \ref{fig:architecture}.

The input to our framework at time $t$ is a stream of frames $\mathcal{I}_t=\{I_{0}, I_{1} \dots,  I_{t-1}, I_t\}$ and the output is $\mathcal{O}_t = \{O_{0}, O_{1}, \dots, O_{t-1}, O_t\}$, where $\mathcal{O}_t = ({P}_t,{C}_t)$.

${P}_t \in [0,1]$ is the GPM prediction for a particular frame at time $t$. ${C}_t$ is the predicted gesture category at time $t$ where ${C}_{t} = \{1,2,\dots, N+1\}$, where $N$ is the number of gesture classes and one for the no-gesture class.

\subsection{Spatiotemporal Encoder \label{encoder}} The Spatiotemporal encoder forms the backbone of our architecture. The goal of this module is to extract rich spatial and temporal features from the raw input video frames suitable for gesture recognition. The resulting features encode both the appearance and motion information present in the video frames. The encoder maps the current frame ${I}_{t}$ to the spatiotemporal features: ${\Phi}_{enc} : \mathcal{I} \rightarrow \mathcal{F}$ where $\mathcal{F} \in \mathbb{R}^{d}$ and $\mathcal{I} = \{I_{1},\dots, I_{t}\}$. ${d}$ is the dimension of the feature map.

3DCNNs have shown promise in extracting local and short term spatiotemporal features from a sequence of frames \cite{3dcnn},~\cite{3dcnn_benefits_1}. Therefore, we leverage a 3DCNN network for extracting features from the raw video frames. Spatial max pooling is used to reduce the spatial resolution but the features are not pooled temporally for maintaining frame level granularity. The output of the 3DCNN is connected to two linear layers.

While 3DCNNs are effective in modelling the short term dependencies, recurrent units like Gated Recurrent Unit (GRU) have been proposed for capturing the long term dependencies by \cite{gru}. The output $\mathcal{G} \in \mathbb{R}^{d}$ of the 3DCNN network is fed to GRU. GRU takes the features from the 3DCNN network and the hidden state representation from the previous time step as the input and outputs the features for the current video frame. We use the following formulation for the GRU: 

\begin{equation}
    \begin{split}
        z_t & = \sigma(g_tU^z + f_{t-1}W^z) \\
        r_t & = \sigma(g_tU^r + f_{t-1}W^r) \\
        s_t & = tanh(g_tU^h + (f_{t-1}\odot r_t  )W^h) \\
        f_t & = (1-z_t)\odot s_t+ z_t \odot f_{t-1}     
    \end{split}
\end{equation}

where, $g_t$ are the features extracted from the 3DCNN and $f_t$ is the output of the encoder network. $z_t$ and $r_t$ represent the output of update and reset gates at time $t$ respectively and $U,W$ are the learned parameters. These features are used as input to the GPM and classification modules as explained in the next sections.

\subsection{Gesture Progression Modelling (GPM)\label{gpm}}
The Gesture Progression Modelling (GPM) module models the temporal progression of the gestures at frame level granularity. It regresses the feature embedding into the progression value.
${\Phi}_{gpm} : \mathcal{F} \rightarrow \mathcal{P}$ where $\mathcal{P} \in [0,1]$. 
In this work, we use the elapsed duration as a measure of gesture progression. The elapsed duration is normalized by the duration of the gesture to accommodate gestures of varying length. If a gesture starts at frame $t_s$ and ends at frame $t_e$, the GPM value $\Phi_{gpm_t}$ at time $t$ is defined as:
\begin{equation}
    \Phi_{gpm_t} = 
    \begin{cases}
        \frac{t - t_s}{t_e - t_s}, & \text{if } {t_s \le t  \le t_e} \\
        0,                         & \text{otherwise}    
    \end{cases}
\end{equation}

{$\Phi_{gpm_t}$} is set to zero for background frames.
This module enables our framework to do reliable early gesture detection as it predicts the completion ratio as the gesture moves towards completion. Our method allows the flexibility to configure different stages of prediction for every gesture. It also provides the option of modifying the gesture trigger points even after the model is trained. This saves gesture re-annotation as well as model retraining efforts.

\subsection{Gesture Classification}
The objective of the classification module is to identify the category of the gesture. The module is expected to distinguish among the gestures as well as the no-gesture class. To model this, we add an extra category representing the no-gesture to the existing gesture categories. 

${\Phi}_{class} : {\mathcal{F}} \rightarrow {C}$ where ${C}_{t} = \{1,2,\dots, N+1\}$ where $N$ is the number of gesture classes and one for the no-gesture class. We train this module with a weighted cross entropy loss $\mathcal{L}_{class}$ to balance the learning between gesture classes and no-gesture class, where the weights are inversely proportional to the number of class samples.

\subsection{Loss function} We jointly train the GPM branch with a mean square loss, $\mathcal{L}_{gpm}$ and the classification branch with the weighted cross entropy loss, $\mathcal{L}_{class}$. $\mathcal{L}_{gpm}$ is defined as:
\begin{equation}
    \mathcal{L}_{gpm} = \frac{1}{T}\sum_t (p_t - \hat{p_t})^2
\end{equation}
where, $p_t$ and $\hat{p_t}$ are the predicted and ground truth gesture progression values at time $t$. $\mathcal{L}_{class}$ is defined as:

\begin{equation}
    \mathcal{L}_{class} = -\frac{1}{T}\sum_t w_t \log c_t
\end{equation}
The weights $w_t$ are inversely proportional to the frequency of the gesture category at time $t$. $c_t$ is the predicted probability corresponding to the ground truth category. The final objective for training the network is given by:
\begin{equation}
    \mathcal{L} = \mathcal{L}_{gpm} + \lambda \mathcal{L}_{class}
\end{equation}
where, $\lambda$ is the hyper-parameter for weighting the respective losses.

\begin{figure*}[h]
\begin{center}
\includegraphics[width=\linewidth]{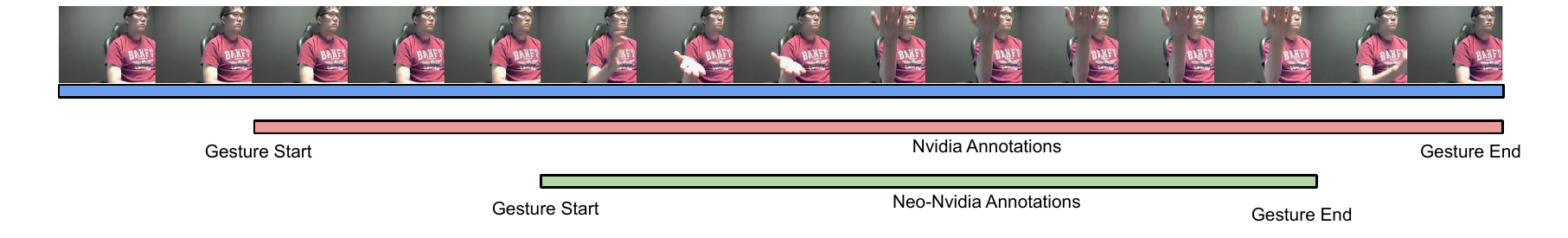}
\end{center}
\caption{The Neo-Nvidia annotations accurately localize the start and end frames of the gestures. In this figure, we show the annotations for an instance of "swipe-up" gesture performed by a subject. Unlike the existing annotations, the Neo-Nvidia annotations are strongly segmented and do not contain the background or no-gesture frames.}
\label{fig:neo_annotation}
\end{figure*}

\subsection{Gesture Inference}
The inference strategy of our method is based on the level of gesture progression. In \textit{offline} setting, the gesture prediction is triggered at the peak of the predicted GPM curve which also represents that the gesture has completed. For \textit{online} setting, we trigger gesture detection when the GPM output exceeds a predefined threshold. In both the settings, the probability vector corresponding to the detected location is used for the classification of the gesture.

\section{Implementation Details}

\subsection{Architecture}
The Spatiotemporal encoder consists of a 3DCNN and GRU network to extract spatiotemporal features from the raw video frames as shown in Figure~\ref{fig:spatio_temporal_architecture}. Conv3D represents a 3D convolution layer with kernel size 3 x 3 x 3 and denoted feature maps, Linear(n) is a fully connected linear layer of \textit{n} output units and MaxPool3D is a max pooling layer. All the 3D convolution layers are followed by batch normalization~\cite{batchnorm} to speedup training. All the max pooling layers use the kernel size of 1 x 2 x 2 to retain temporal resolution. ReLU is used as the activation function after every convolutional and linear layer. The output of the 3DCNN is connected to a GRU with 1024 units. 

The GPM Module consists of a linear layer followed by a sigmoid activation to constrain the outputs between 0 and 1 and is trained with a mean square loss $\mathcal{L}_{gpm}$. The gesture classification module consists of a single linear layer with softmax activation and outputs the class probabilities for each frame.

\subsection{Training}
Each video is subsampled into 80 frames by using nearest neighbour sampling and resized to 160 x 120 spatial dimensions. We crop the frames to 112 x 112 at random spatial position during training and use the center crop during inference. We use stochastic gradient descent with a learning rate of 0.001 and reduce it by a factor of 10 after every 100 epochs. Momentum of 0.9 and weight decay of 0.005 is used. We observed that the value of $1.0$ for weighting the losses between the two branches works best for having a balanced learning among the branches. To avoid overfitting, we randomly perturb the video frames with spatial rotation ($\pm{25}^\circ$), spatial scaling ($\pm{20}$\%), temporal scaling ($\pm{20}$\%), non-linear temporal scaling and temporal translation ($\pm{5}$ frames). For every video, a random value is sampled from a uniform distribution with these intervals. We further avoid overfitting by following every 3D convolution with a volumetric dropout~\cite{tompson2015efficient} and linear layers by linear dropout~\cite{hinton2012improving}. Volumetric dropout helps the model by promoting independence between feature maps. The dropout probabilities are set at 0.1 for 3DCNN layers and 0.85 for linear layers. We also clip the gradients (-10,10) to avoid gradient explosion ~\cite{pascanu2013difficulty}.
Our framework is written in torch7 \cite{collobert2011torch7} and trained on NVIDIA Titan X GPUs. 

We trained our model on depth modality first and used it to initialize weights for other modalities after appropriate inflation.  For example,  we inflated the first layer from 1 channel to 3 channels for RGB modality and 1 channel to 2 channels for optical flow. While inflating to color, we divide the weights by 3 and for flow we divide them by 2 for  normalizing the activations.

\section{NVIDIA Dataset}
\subsection{Dataset}
The NVIDIA gesture dataset consists of dynamic hand gestures collected in a car simulator under different lighting conditions. The gesture categories are focussed towards designing human computer  interaction (HCI) interfaces which makes it an important benchmark for online gesture analysis. The dataset consists of 25 dynamic gesture classes like hand and finger swipes in different directions, pointing index finger, moving two finger in the clockwise or anti clockwise direction etc. A total of 20 subjects participated in the data collection campaign, resulting in a dataset of 1532 video samples. The dataset has multiple modalities: Color, Depth and pair of IR streams and is split into 1050 training videos and 482 test videos. We perform extensive experiments on this dataset and compare our framework with the baseline and current state-of-the-art on this dataset~\cite{pavloctc}. 

\subsection{Neo-NVIDIA Annotations}
 The videos of the NVIDIA dataset are weakly annotated as the annotated start and end frames also contain the background or no-gesture frames. This makes it unsuitable for our approach and in general gesture localization tasks. To overcome this, we annotated the NVIDIA gesture dataset with exact gesture start and end boundaries. The frame at which the subject begins to execute the gesture is marked as the starting frame and the frame at which gesture is completed is marked as the end frame as shown in Figure~\ref{fig:neo_annotation}. A team of experienced annotators annotated the dataset by observing the depth and color videos. Every video was annotated and reviewed by multiple annotators to maintain the quality of annotation. We will release these new annotations \footnote{https://github.com/vguptai/Neo-Nvidia-Annotations} for advancing the research in this domain.

\begin{table}
\begin{center}
\begin{tabular}{|l| cc | cc | cc |}
\hline
& \multicolumn{6}{c|}{NTtD}\\
\cline{2-7}
 & \multicolumn{2}{c}{0.25} &  \multicolumn{2}{|c|}{0.50} & \multicolumn{2}{c|}{0.75}\\
\hline
Modality & FPR & TPR & FPR & TPR & FPR & TPR\\
\hline\hline
Depth & 6.9 & 89.5 & 1.7 & 49.5 & 0.3 & 12.5\\
Color & 11.3 & 91.2 & 1.8 & 43.6 & 0.3 & 7.9\\
Flow & 11.1 & 92.5 & 1.9 & 45.6 & 0.2 & 7.1\\
IR & 11.6 & 84.3 & 1.4 & 33.7 & 0.2 & 6.1\\
\hline
\end{tabular}
\end{center}
\caption{True Positive Rate (TPR) and False Positive Rate (FPR) across different Normalized Time to Detect (NTtD) values on the NVIDIA dataset for different modalities.}
\label{table:early_prediction_result}
\end{table}

\begin{table}
\begin{center}
\begin{tabular}{|l|c|c|c|c|c|}
\hline
Modality & Depth & Flow & Color & IR & Fusion\\
\hline\hline
AUC & 95.1 & 94.3 & 92.9 & 89.3 & \textbf{95.2}\\
\hline
\end{tabular}
\end{center}
\caption{Area under the Curve (AUC) of the Receiver Operating Characteristics (ROC)  curve on different modalities on the NVIDIA dataset. "Fusion" represents AUC after fusing all the modalities.}
\label{table:online_results_nvidia}
\end{table}

\begin{figure}[t]
\centering
\includegraphics[width=0.75\columnwidth]{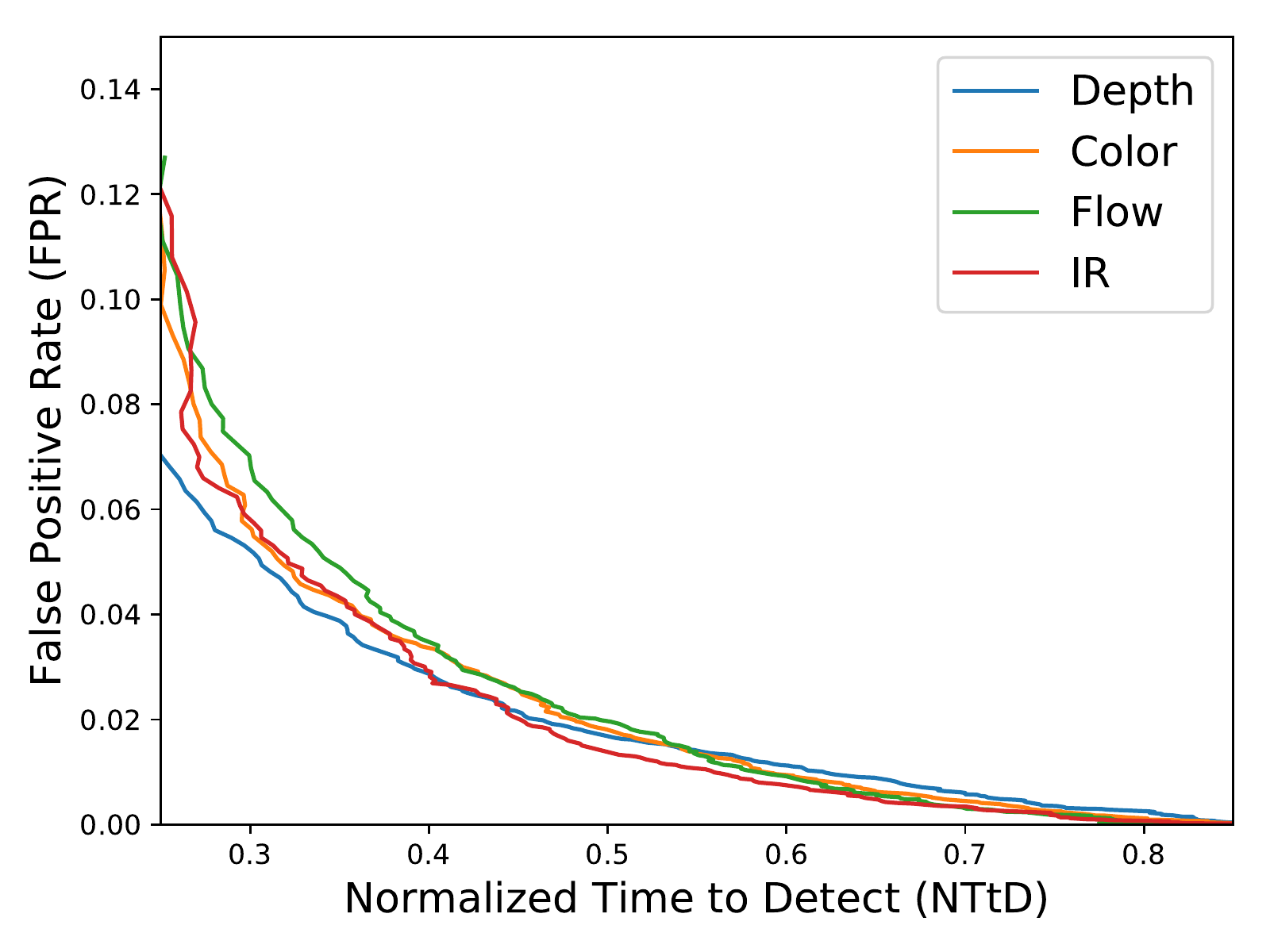}
\caption{Normalized Time to Detect (NTtD) vs False Positive Rate (FPR) on the NVIDIA dataset for depth, color, flow and IR modality.}
\label{fig:nttd_pavlo}
\end{figure}
\section{NVIDIA Dataset Experiments}
\subsection{Early and Online Gesture Recognition}
In \textit{online setting}, the framework processes an incoming stream of frames and outputs the classification and progression predictions for each frame. We select a detection threshold value $\epsilon \in [0,1]$ and when the GPM output exceeds the selected threshold, the class probability scores of the corresponding frame are used to determine the predicted gesture class. We compute the Normalized Time To Detect (NTtD)~\cite{early_prediction_2} to measure the performance of our system for early prediction. NTtD is defined as the ratio of event duration that the detector observes before the event prediction. We report the False Positive Rate (FPR) and True Positive Rate (TPR) across different mean NTtD values for the correctly recognized gestures. TPR is defined as the ratio of correctly predicted gesture frames to the total gesture frames and FPR is the ratio of incorrectly classified gesture frames to the no-gesture frames. 

Our framework is able to recognize gestures by processing only 25\% of their duration with a low FPR and high TPR as shown in Table~\ref{table:early_prediction_result}. In Figure~\ref{fig:nttd_pavlo}, we plot the detailed FPR vs NTtD characteristics and observe that the FPR is inversely proportional to the NTtD, which is expected as the model becomes more confident on observing longer durations of the gesture. 

To analyze the detection performance of our system across different detection thresholds, we also plot the receiver operating characteristic (ROC) curve~\cite{early_prediction_2} of TPR and FPR at different threshold values and report the area under the curve (AUC) for different modalities and their fusion in Table \ref{table:online_results_nvidia}.

\begin{table}[t]
\begin{center}
\begin{tabular}{|l|c|c|}
\hline
Modality  & Ours  & Molchanov~\textit{et al.} \\
\hline\hline
IR & 68.7 & 63.5 \\
Color & 75.9  & 74.1     \\
Flow & 78.2  & 77.8      \\
Depth & 85.5  & 80.3     \\
IR Disparity (ID)  & - & 57.8 \\
\hline
Flow + Color & 80.3 & 79.3 \\
Depth + Flow & 85.5 & 82.4 \\
Depth + Color & 86.1 & - \\
Depth + Color + Flow & 86.3 & 81.5 \\
Depth + Color + Flow + IR & \textbf{87.8}  & 83.4 \\
Depth + Color + Flow + IR + ID& -  & \textbf{83.8} \\
\hline
\hline
Human Accuracy & & \textbf{88.4} \\
\hline
\end{tabular}
\end{center}
\caption{Comparison of~\textit{Offline Classification} accuracy (\%) of the proposed method with~\cite{pavloctc} for different modalities and their fusion on the NVIDIA gesture dataset.~\cite{pavloctc} report a human accuracy of 88.4\% on this dataset.}
\label{table:offline_results_nvidia}
\vspace{-1em}
\end{table}

\begin{figure*}[h]
    \centering
    \includegraphics[width=\textwidth, height=25em]{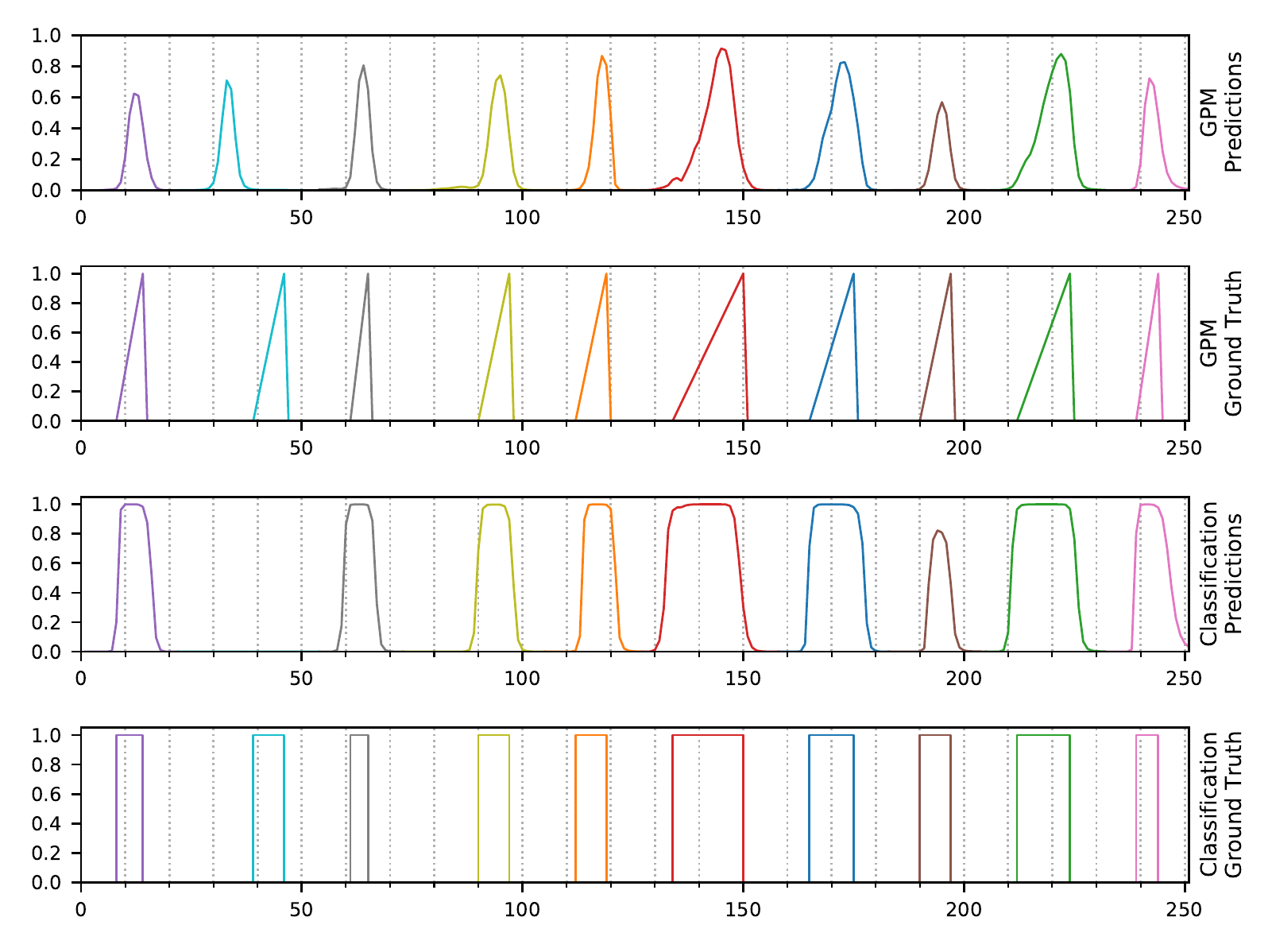}
    \caption{Plot of the model predictions vs time for gestures in the test set. The first two rows correspond to the Gesture Progression Module and the last two are for the classification branch. The second peak is an example of a failure case in which the GPM and classification module fail to model the progression of the gesture.}
    \label{gesture_plot}
\end{figure*}

\subsection{Offline Gesture Recognition}
In Table~\ref{table:offline_results_nvidia}, we compare the offline performance of our method with~\cite{pavloctc} for different modalities and combinations. Our method achieves state-of-the-art accuracy on the NVIDIA dataset and further approaches human level accuracy. Fig.~\ref{gesture_plot} depicts the ground truths and predictions of the GPM and classification module. We can observe that both the GPM and classification modules trigger at similar time frames for the successful cases and fail to align in the failure cases (second peak in the plot). In our experiments, we observe that depth modality outperforms other modalities, which can be explained by the fact that depth data is less sensitive to ambient conditions like lighting, background noise, etc. We use a simple weighted average strategy over the conditional probabilities to combine the predictions of different input modalities. The ensemble weights were estimated through a linear classifier trained on training data.

\begin{table}[t]
\begin{center}
\begin{tabular}{|l|c|}
\hline
Modality  & Jaccard Index \\
\hline\hline
Depth & 0.60     \\
Flow  & 0.54     \\
Color & 0.53     \\
IR & 0.47 \\
Depth + Color + Flow + IR & 0.61 \\
\hline
\end{tabular}
\end{center}
\caption{Localization results on the NVIDIA dataset. The Jaccard index indicates the mean overlap between predictions and the ground truth across gesture categories.}
\label{table:offline_iou_nvidia}
\vspace{-1em}
\end{table}

\subsection{Gesture Localization}
Our tightly segmented Neo-NVIDIA annotations also allow us to perform gesture localization on this dataset. We provide a benchmark localization performance in Table \ref{table:offline_iou_nvidia}. To the best of our knowledge, we are the first to do this on the NVIDIA dataset.
We compute the Intersection over Union (IOU) of gesture detections and ground truth and report the mean Jaccard Index. Jaccard Index is the standard metric for localization task and has been used by~\cite{pigou2018beyond},~\cite{pavloctc}.

\subsection{Ablation Studies}

\subsubsection{Spatiotemporal Encoder Architecture}
We evaluate the contribution of 3D convolutions and recurrent units in Table \ref{table:arch_component_analysis} and observe deterioration in the performance by using a linear aggregator of 3DCNN features when compared with GRU based recurrent units. This is expected since the linear network can not model long term temporal information. We further evaluate the architecture in which we use 2DCNN as the feature extractor and model temporal information using a GRU network and observe a decrease in the classification accuracy. From this analysis, we conclude that both 3DCNN and GRU components, are independently crucial to the state-of-the-art performance of our network.

\begin{table}[t]
\begin{center}
\setlength{\tabcolsep}{3pt}
\begin{tabular}{|l|c|c|c|}
\hline
Architecture & 2DCNN-GRU & 3DCNN-Linear & 3DCNN-GRU\\
\hline\hline
Acc (\%) & 77.4 & 81.5 & 85.5\\
\hline
\end{tabular}
\end{center}
\caption{~\textit{Offline Classification} accuracy(\%) of our approach under different architecture settings of the Spatiotemporal Encoder. Results are reported on depth modality.}
\label{table:arch_component_analysis}
\vspace{-1em}
\end{table}

\subsubsection{Gesture Progression Modeling}
In Table \ref{table:gpm_analysis}, we study the efficacy of the GPM in detecting the gesture correctly and at the correct location. We define $\mathcal{I}$ as the consensus set of frames which participate in voting for the final category prediction. In the baseline setting, GPM branch is not used and global voting is done~\cite{pigou2018beyond}. In this setting, the consensus set is $\mathcal{I} = \{I_1, I_2, \dots, I_N\}$ where $N$ is the number of frames in the video. In the next settings, we use the GPM branch to choose the consensus set for classification at different thresholds. Formally, the consensus set $\mathcal{I} = \{I_i:  \Phi_{gpm_i} > \tau m\}$ where $\tau$ is our ratio and $m = max\{\Phi_{gpm_i}, i = 1, 2 \dots N$\} is the maximum gesture progression level predicted by the GPM. We include results with various $\tau$ values in Table \ref{table:gpm_analysis}. We observe the best accuracy when using 100\% progression level which is identical to selecting the frame with maximum progression value. This analysis shows that the GPM branch is able to accurately predict the completion of the gesture within the gesture ground truth. An explanation for inferior performance of simple voting is that it can not handle the false positives caused due to unintentional hand movements.  In noisy videos, such false positives can dominate the actual gesture frames. GPM solves this problem by allowing the model to focus on relevant gesture frames.

\begin{table}[t]
\begin{center}
\begin{tabular}{|l|c|c|}
\hline
Threshold  & Depth& Color\\
\hline\hline
Baseline (Global Voting)       & 84.2  & 74.7 \\
GPM @ 75\% & 84.7 & 75.9    \\
GPM @ 85\% & 84.9 & 75.5     \\
GPM @ 95\% & 84.9 & 75.5     \\
GPM @ 100\%  & 85.5 & 75.9     \\
\hline
\end{tabular}
\end{center}
\caption{~\textit{Offline Classification} accuracy (\%) at different threshold ratios to the maximum GPM prediction in the video for depth and color modality. In~\textit{Baseline} setting, the GPM branch is not used and global voting is performed for classification.}
\label{table:gpm_analysis}
\end{table}

\begin{table}
\begin{center}
\begin{tabular}{|l|l|r|}
\hline
Approach & Modality & Acc \\ 
\hline\hline
\cite{pigou2018beyond}  & Color+Depth+Skeleton & 97.2  \\
\cite{pavloctc} & Color+Depth+Flow & 98.2 \\
Ours & Depth & 95.3 \\
Ours & Color & 96.8  \\
Ours & Flow & 94.6 \\ 
Ours & Color+Depth+Flow & 97.7 \\
\hline
\end{tabular}
\end{center}
\caption{Results on Montalbano dataset. Comparison of~\textit{Offline Classification} accuracy (\%) of the proposed method with the state-of-the-art on pre-segmented videos for different modalities and fusion.} 
\label{table:chalearn_results}
\end{table}

\section{Montalbano Dataset Experiments}\label{ssec:chalearn}
\subsection{Dataset}
The Montalbano dataset \cite{chalearn2014dataset} is a large dataset of around 14K gestures belonging to 20 categories and performed by 27 subjects under varying conditions. The videos were collected using Microsoft Kinect and have color, depth and skeletal information. Multiple gestures are present in each video and for each gesture, along with the gesture category, the start and end frame have also been annotated.  
We conduct experiments on the Montalbano dataset to comprehensively compare our method with the early gesture detection baseline~\cite{pavloctc} and demonstrate that our method achieves competitive results on this dataset also.

\begin{table}[t]
\begin{center}
\begin{tabular}{|l|c|}
\hline
Modality  & Jaccard Index \\
\hline\hline
Depth & 0.89     \\
Flow  & 0.87     \\
Color & 0.90     \\
Depth + Color + Flow & 0.91 \\
\hline
\end{tabular}
\end{center}
\caption{Localization results on the Montalbano dataset. The Jaccard index indicates the mean overlap between predictions and the ground truth across gesture categories.}
\label{table:offline_iou_chalearn}
\end{table}

\subsection{Experimental Results}

For early gesture recognition, we report TPR and FPR of 83\% and 5.6\% respectively for the NTtD of 20\% on color modality. For measuring the offline gesture recognition performance, we compare our results with~\cite{pigou2018beyond},~\cite{pavloctc} in Table \ref{table:chalearn_results} across different modalities and achieve competitive results. We also measure the localization performance of our method and report Jaccard Index as per Table \ref{table:offline_iou_chalearn} for all the modalities and achieve comparable result of 0.91 as reported by \cite{pigou2018beyond}.

\section{Conclusion}
Early and online detection of gestures is important for designing responsive and real time gesture based interfaces. In this work, we proposed a multitask learning framework that models the progression of the gesture (GPM) along with frame level classification for performing early gesture detection. The proposed framework works well on both online and offline settings. In \textit{online setting}, our method is able to detect gestures before completion with high True Positive Rate (TPR) and low False Positive Rate (FPR). For \textit{offline} gesture detection, we outperform the state-of-the-art accuracy on the NVIDIA dataset and report competitive results on Montalbano dataset. To further the research, we contribute a new set of tightly segmented annotations for the NVIDIA dataset and setup a new localization baseline.
\\
\\
\textbf{Acknowledgement:} We would like to thank Shuaib Ahmed, Mallikarjun BR, Neha Tarigopula and Sanath Narayan for their valuable feedbacks and discussions. We gratefully acknowledge Brijesh Pillai and Partha Bhattacharya at Mercedes-Benz Research and Development India, Bangalore
for providing the funding and infrastructure for this work.

\end{document}